\documentclass[journal]{IEEEtran}

\usepackage{cite}
\usepackage[pdftex]{graphicx}
\usepackage{mathtools}
\usepackage{amsmath}
\usepackage{array}
\usepackage{mathrsfs}
\usepackage{amssymb}
\usepackage{booktabs}
\usepackage{algorithm}
\usepackage{algpseudocode}
\usepackage{multirow}
\usepackage{tabularx}
\usepackage{tabulary}
\usepackage{dcolumn}
\usepackage{threeparttable}
\usepackage{array}
\usepackage{bbm}
\usepackage{color, soul}

\begin{document}

\title{Smooth Pinball Neural Network for \\Probabilistic Forecasting of Wind Power}

\author{Kostas~Hatalis,~\IEEEmembership{Student Member,~IEEE,}
        Alberto~J.~Lamadrid,~\IEEEmembership{Senior Member,~IEEE,}
        Katya~Scheinberg,
        and~Shalinee~Kishore,~\IEEEmembership{Senior Member,~IEEE}
\thanks{K. Hatalis and S. Kishore are with the Department
of Electrical and Computer Engineering, Lehigh University, Bethlehem,
PA, 18015, USA (e-mail: {\tt kmh511@lehigh.edu}).}
\thanks{K. Scheinberg is with the Department of Industrial Engineering and System Engineering, Lehigh University.}
\thanks{A. J. Lamadrid is with the Department of Economics and the Department of Industrial and System Engineering, Lehigh University, Bethlehem, PA, 18015 USA (e-mail: {\tt ajlamadrid@lehigh.edu})}
}


\maketitle

\begin{abstract}
Uncertainty analysis in the form of probabilistic forecasting can significantly improve decision making processes in the smart power grid for better integrating renewable energy sources such as wind. Whereas point forecasting provides a single expected value, probabilistic forecasts provide more information in the form of quantiles, prediction intervals, or full predictive densities. This paper analyzes the effectiveness of a novel approach for nonparametric probabilistic forecasting of wind power that combines a smooth approximation of the pinball loss function with a neural network architecture and a weighting initialization scheme to prevent the quantile cross over problem. A numerical case study is conducted using publicly available wind data from the Global Energy Forecasting Competition 2014. Multiple quantiles are estimated to form 10\%, to 90\% prediction intervals which are evaluated using a quantile score and reliability measures. Benchmark models such as the persistence and climatology distributions, multiple quantile regression, and support vector quantile regression are used for comparison where results demonstrate the proposed approach leads to improved performance while preventing the problem of overlapping quantile estimates.
\end{abstract}


\IEEEpeerreviewmaketitle


\section{Introduction}

\IEEEPARstart{I}{n} the last thirty years wind power has experienced rapid global growth, and in some countries it is the most used form of renewable energy. However, due to the chaotic nature of weather, variable and uncertain wind power production poses planning and operational challenges unseen in conventional generation. From the grid operator's perspective, uncertainty in wind production could cause inefficiencies in the power flow, operating reserve requirements, stochastic unit commitment, and electricity market settlements. \cite{makarov2009operational,bukhsh2016integrated,botterud2012wind}. From the wind generator's perspective, reliable wind forecasts are needed for several operations at a wind farm, ranging from energy storage control to bidding and trading in energy markets. Thus, to ensure both stable grid operations and continued growth and increased penetration of wind power, highly reliable forecasting of wind power production is needed.

Traditionally wind power prediction has focused on developing deterministic point forecasts which provide an expected output for a given look-ahead time. Forecasting horizons can be categorized into several scales: very short term (several seconds or minutes ahead), short term (several hours to days ahead), long term (weeks or months ahead), and seasonal. However, point forecasting can result in unavoidable errors which can be significant and they also lack information on uncertainty.  Therefore, a large research effort has been conducted recently by the renewables forecasting community \cite{hong2016probabilistic} to produce full probabilistic predictions which derive quantitative information on the associated uncertainty of power output. For example, to capture uncertainty of wind power, forecasting errors can be statistically analyzed and modeled by the Beta distribution. However, such  assumption may not be applicable for short term forecasting and thus researchers are looking at different approaches for probabilistic wind power forecasts by quantifying forecasting uncertainty. Although various methods have been proposed, it is still a challenge to make accurate and robust probabilistic predictions for highly nonlinear and complex data, such as wind. 

Probabilistic wind models are based on either meteorological ensembles that are obtained by a weather model \cite{giebel2003using} or on statistical methods \cite{foley2012current}. Focusing on the statistical approach, these methods can be applied to forecast full predictive distributions in the form of quantiles or prediction intervals (PIs). A very common approach in probabilistic forecasting of wind power is to apply quantile regression (QR) which can be used to estimate different wind power quantiles. The QR model can be solved by linear programming algorithms and many variations have been proposed over the years. In \cite{bremnes2004probabilistic} local QR is applied to estimate different quantiles, while in \cite{nielsen2006using} a spline-based QR is used to estimate quantiles of wind power. In \cite{landry2016probabilistic} quantile loss gradient boosted machines are used to estimate many quantiles and in \cite{juban2016multiple} multiple quantile regression is used to predict a full distribution with optimization achieved by using the alternating direction method of multipliers. Quantile regression forests \cite{juban2008uncertainty} have also been applied in forecasting. These are an extension of regression forests based on classification and regression trees. 

Once a probabilistic forecast is obtained it could be used for several applications related to managing wind farms. For instance, in \cite{doherty2005new} the optimal level of generation reserves is estimated using the uncertainty of wind power predictions, and in \cite{usaola2004benefits,castronuovo2004optimization} the optimization of wind energy production is investigated taking into account the forecasts of a probabilistic prediction method. Additionally, increased revenues can be obtained using bidding strategies built on predictive densities, as shown in \cite{bathurst2002trading,pinson2007trading}. Wind power density forecasting can be used for analysis on probabilistic load flow, as in \cite{karakatsanis1994probabilistic}. Machine learning frameworks have also been used for uncertainty prediction of renewables such as nearest neighbors \cite{mangalova2016k}, neural networks \cite{sideratos2012probabilistic}, and extreme learning machines \cite{wan2014probabilistic}. A thorough overview of probabilistic wind power forecasting is provided in \cite{zhang2014review}. 

Since QR is often used as a comprehensive strategy for providing the conditional distribution of a response $y$ given $x$, several of its parametric and nonparametric variants are highlighted here. In a generalization of the quantile regression model \cite{powell1984least,powell1986censored} introduce the censored quantile regression model, which consistently estimates conditional quantiles when observations on the dependent variable are censored. \cite{yu1998local} propose a nonparametric version of regression quantile estimation by using kernel weighted local linear fitting. Yu and Jones \cite{chen2009copula} propose a copula-based nonlinear quantile autoregression, addressing the possibility of deriving nonlinear parametric models for different conditional quantile functions. Quantile regression can also be hybridized with machine learning methods to form powerful nonlinear models. The idea of support vector regression is introduced for quantile regression model, yielding support vector quantile regression (SVQR) by \cite{hwang2005simple}. SVQR can estimate quantile regression model in non-linear and high dimensional spaces, but it requires solving a quadratic programming problem. 

Due to their flexibility in modeling complex nonlinear data sets, artificial neural networks are another powerful class of machine learning algorithms that have been used to enhance quantile regression. Taylor \cite{taylor2000quantile} proposes a quantile regression neural network (QRNN) method, combining the advantages of both quantile regression and a neural network. This method can reveal the conditional distribution of the response variable and can also model the nonlinearity of different systems. The author applies this method to estimate the conditional distribution of multi-period returns in financial systems, which avoids the need to specify explicitly the explanatory variables. However, the paper does not address how the network was optimized. The same QRNN was later used by \cite{feng2010robust} for credit portfolio data analysis where results showed that QRNN is more robust in fitting outliers compared to both local linear regression and spline regression. In \cite{xu2016quantile} an autoregressive version of QRNN is used for applications to evaluating value at risk and \cite{cannon2011quantile} implements the QRNN model in R as a statistical package. 

In all the nonlinear QR approaches mentioned, estimation of each quantile is conducted independently. In the case of estimating multiple quantiles, this could lead to what is known as the quantile cross over problem, where a lower quantile overlaps a higher quantile.  Or equivalently, a prediction interval for a lower probability (e.g., range in which 10\% of future values are predicted to lie) exceeds that of a higher probability (e.g., the range in which 20\% of the future values are predicted to lie). This is undesirable as it violates the principle of cumulative distribution functions where their associated inverse functions should be monotonically increasing. A possible way to prevent this issue is to utilize simple heuristics of reordering estimated quantiles. However this approach does not have a strong theoretical foundation and may lead to inappropriate quantiles. The solution then is to optimize quantiles together with non-crossing constraints. In \cite{takeuchi2004non,takeuchi2006nonparametric} a constrained support vector quantile regression (CSVQR) method was developed with non-crossing constraints where it was used to fit quantiles on static data. However, CSVQR is computationally very expensive and slow to train. In order to address the issue of the quantile crossover problem, in addition to the problem of dealing with nonlinearity in wind forecasting, we propose a novel model which we call the smooth pinball neural network (SPNN). The main contributions of our approach can be summarized as follows:
\begin{enumerate}
	\item We use a new and simple objective function which is a logistic based smooth approximation of the pinball loss function for multiple quantile regression and show how to apply it to train a neural network with standard gradient descent back-propagation.
	\item We introduce an initialization weighting scheme to prevent the quantile cross over problem and improve estimation.
	\item We are the first to showcase a quantile based neural network for probabilistic forecasting of wind with a sliding window of training data.
	\item We design experiments to validate our model using publicly available data from 10 wind farms from the Global Energy Forecasting Competition 2014 and benchmark performance with common and advanced methods.
\end{enumerate}

The contents of the paper are: in Section \ref{s:prob} we provide the mathematical background on probabilistic forecasting, quantile regression, and evaluation methods. In Section \ref{s:mod} we go over our model, its architecture, training, and weighting initialization scheme. Results and discussion of our case study are presented in Section \ref{s:res}.  We conclude the paper and present future research directions in Section \ref{s:con}.

\section{Probabilistic Forecasting}\label{s:prob}

This section highlights the underlying mathematics in probabilistic forecasting, overviews linear quantile regression, and summarizes the main evaluation methods for quantile forecasts. Given a random variable $Y_t$ such as wind power at time $ t $, its probability density function is defined as $f_t$ and its the cumulative distribution function as $F_t$. If $ F_{t} $ is strictly increasing, the quantile $ q_{t}^{(\tau)} $ of the random variable $ Y_{t} $ is uniquely defined as the value $ x $ such that $ P(Y_t < x) = \tau $ or equivalently as the inverse of the distribution function $ q_{t}^{(\tau)}  = F_{t}^{-1}(\tau) $. A quantile forecast  $ \hat{q}_{t+z}^{(\tau)} $ with nominal proportion $ \tau $ is an estimate of the true quantile $  q_{t+z}^{(\tau)} $ for the lead time $ t+z $, given a predictor values (such as numerical wind speed forecasts). Prediction intervals then give a range of possible values within which an observed value is expected to lie with a certain probability $ \beta \in [0,1] $. A prediction interval $ \hat{I}^{(\beta)}_{t+z} $ produced at time $ t $ for future horizon $ t + z $ is defined by its lower and upper bounds, which are the quantile forecasts $ \hat{I}^{(\beta)}_{t+z} = \left[ \hat{q}_{t+z}^{(\tau_{l})} ,\hat{q}_{t+z}^{(\tau_{u})} \right] = \left[ l_{t}^{(\beta)},u_{t}^{(\beta)} \right] $ whose nominal proportions $ \tau_l $ and $ \tau_u $ are such that $ \tau_u - \tau_l = 1-\beta $. 

In probabilistic forecasting, we are trying to predict one of two classes of density functions, either parametric or nonparametric. When the future density function is assumed to take a certain distribution, such as the Normal distribution, then this is called parametric probabilistic forecasting. For a nonlinear and bounded process such as wind generation, probability distributions of future wind power for instance may be skewed and heavy-tailed distributed \cite{dorvlo2002estimating}. Else if no assumption is made about the shape of the distribution, a nonparametric probabilistic forecast $ \hat{f}_{t+z} $ \cite{pinson2007non} can be made of the density function by gathering a set of $ M $ quantiles forecasts such that $ \hat{f}_{t+z} = \left\lbrace  \hat{q}_{t+z}^{(\tau_{m})} ,m=1,...,M|0\leq \tau_1 <  ... < \tau_M \leq 1 \right\rbrace  $ with chosen nominal proportions spread on the unit interval. In this paper, we consider forecasting wind power on the resolution of one hour (predicting outwards to a month worth of values). On this resolution scale of an hour, the wind density may fluctuate therefore making nonparametric forecasting more ideal then fitting a parametric density \cite{zhang2014review}.

Quantile regression is a popular approach for nonparametric probabilistic forecasting. It was introduced by \cite{koenker1978regression} for estimating conditional quantiles and is closely related to models for the conditional median \cite{koenker2005quantile}. Minimizing the mean absolute function leads to an estimate of the conditional median of a prediction. By applying asymmetric weights to errors through a tilted transformation of the absolute value function, we can compute the conditional quantiles of a predictive distribution. The preferred transformation function is the pin ball loss function as defined by
\begin{equation} \label{pinball}
\rho_{\tau}(u) = \left\lbrace 
\begin{array}{cl}
\tau u    & \mbox{if } u \geq 0 \\
(\tau-1)u & \mbox{if } u < 0
\end{array} \right.,
\end{equation}

\noindent where $ 0 < \tau < 1 $ is the tilting parameter. To better understand the pinball loss, we look at an example for estimating a single quantile. If an estimate falls above a reported quantile, such as the 0.05-quantile, the loss is its distance from the estimate multiplied by its probability of 0.05. Otherwise, the loss is its distance from the realization multiplied by one minus its probability (0.95 in the case of the 0.05-quantile). The pinball loss function penalizes low-probability quantiles more for overestimation than for underestimation and vice versa in the case of high-probability quantiles. Given a vector of predictors $ X_t $ where $ t = 1,...,N $, a vector of weights $ W $ and intercept $ b $ coefficient in a linear regression fashion, the conditional $ \tau $ quantile $ \hat{q}_\tau $ is given by $ \hat{q}_{t}^{(\tau)} = W^\top X_t +b $. The weights and intercept can be estimated by solving the following minimization problem.
\begin{equation} \label{eq3}
\min_{W,b} \frac{1}{N} \sum_{t=1}^{N} \rho_\tau (y_t-\hat{q}_{t}^{(\tau)}),
\end{equation}

\noindent where $ y_t $ is the observed value of the predictand. The problem in Eq. (\ref{eq3}) can be minimized by a linear program. 

\subsection{Evaluation Methods}

In probabilistic forecasting it is important to evaluate the quantile estimates and if desired also evaluate derived predictive intervals. Therefore, we use as evaluation measures the quantile score, interval reliability and sharpness. To evaluate quantile estimates one can use the pinball function directly as an assessment called the quantile score (QS). We choose QS as our main  evaluation measure for the following reasons. When averaged across many quantiles it can evaluate full predictive densities; it is found to be a proper scoring rule \cite{grushka2016quantile}; it is related to the continuous rank probability score; and it is also the main evaluation criteria in the 2014 Global Energy Forecasting Competition (GEFCOM 2014), the source of our testing data. QS calculated over all $ N $ test observations and $ M $ quantiles is defined as
\begin{equation*}
QS = \sum_{t=1}^{N} \sum_{m=1}^{M} \rho_{\tau_m} (y_t - \hat{q}_{t}^{(\tau_{m})})
\end{equation*}

\noindent where $ y_t $ is an observation used to forecast evaluation such future wind power observations. To evaluate full predictive densities, this score is averaged across all target quantiles for all look ahead time steps using equal weights. A lower QS indicates a better forecast. In some applications, it may be needed to have wind forecasts in the form of prediction intervals (PIs) and as such we look at two secondary evaluation measures: reliability and sharpness. Reliability is a measure which states that over an evaluation set the observed and nominal probabilities should be as close as possible and the empirical coverage should ideally equal the preassigned probability. Sharpness is a measure of the width of prediction intervals, defined as the difference between the upper $ u_{t}^{\beta_i} $ and lower $ l_{t}^{\beta_i} $ interval values.   

For measuring reliability, PIs show where future wind power observations are expected to lie, with an assigned probability termed as the PI nominal confidence (PINC) $ 100(1-\beta_i)\% $. Here $ i = 1...M/2 $ indicates a specific coverage level. The coverage probability of estimated PIs are expected to eventually reach a nominal level of confidence over the test data. A measure for reliability which shows target coverage of the PIs is the PI coverage probability (PICP) which is defined by
\begin{equation*}
PICP_i=\frac{1}{N}\sum_{t=1}^{N} \mathbbm{1}\{y_t \in I_{t}^{\beta_i}({x}_t)\}
\end{equation*}

For reliable PIs, the examined PICP should be close to its corresponding PINC. A related and easier to visualize assessment index is the average coverage error (ACE) which is defined by
\begin{equation*}
ACE = \sum_{\ i = 1}^{M/2} |PICP_i-100(1-\beta_i)|.
\end{equation*}

\noindent This assumes calculation across all test data and coverage levels. To ensure PIs with high reliability, the ACE should be as close to zero as possible. A high reliability can be easily achieved by increasing or decreasing the distance between the lower and upper bounds. Thus, the width of a PI can also influence its quality. This is also known as sharpness. PIs that are sharper, i.e.,  narrower, have a higher quality. The interval score (IS) can be used to evaluate the overall skill of PIs that involves sharpness \cite{gneiting2007strictly}. The IS when evaluated with all test data and coverage levels is defined by
\begin{equation*} 
\begin{multlined}
IS = \frac{2}{NM} \sum_{t = 1}^{N} \sum_{\ i= 1}^{M/2}
(u_{t}^{\beta_i} - l_{t}^{\beta_i}) +\\ \frac{2}{\beta_i}(l_{t}^{\beta_i}-y_t) \mathbbm{1}\{y_t < l_{t}^{\beta_i}\} +  \frac{2}{\beta_i}(y_t - u_{t}^{\beta_i}) \mathbbm{1}\{y_t > u_{t}^{\beta_i}\}
\end{multlined}.
\end{equation*}

\noindent The prediction model is rewarded for narrow PIs and is penalized if the observation misses the interval. The size of the penalty depends on $ \beta_i $. Including all aspects of PI evaluation, the IS can be used to compare the overall skill and sharpness of interval forecasts. However, IS cannot identify the contributions of reliability and sharpness to the overall skill. Thus, ACE and IS are both used for evaluation of PIs along with QS for evaluation of quantile estimation.

\section{Smooth Pinball Network Model}\label{s:mod}

We propose to use artificial neural networks for probabilistic forecasting due to their flexibility and strength in dealing with nonlinear data. We can use the pinball function in the objective function of such a neural network to predict quantiles. However, the pinball function $ \rho $ employed by the original linear quantile regression model in Eq. \eqref{pinball} is not differentiable at the origin, $ x = 0 $. The non-differentiability of $ \rho $ makes it difficult to apply gradient based optimization methods in fitting the quantile regression model. Gradient based methods are preferred for training neural networks since they are time efficient, easy to implement, and yield a local optimum. Therefore, we need a smooth approximation of the pinball function that allows for the direct application of gradient based optimization. We call our new model the smooth pinball neural network (SPNN). SPNN is further enhanced to estimate multiple quantiles and has an initialization scheme to prevent the quantile cross-over problem.

Chen et al. \cite{chen1996class} introduced a class of smooth functions for nonlinear optimization problems and applied this idea to support vector machines \cite{lee2001ssvm}. Emulating the work of Chen, a study by Zheng \cite{zheng2011gradient} proposes to approximate the pinball loss function by a smooth function on which a gradient descent algorithm for quantile regression can be applied. They called the resulting algorithm the gradient descent smooth quantile regression model. We extend that model here for the case of a neural network.

\subsection{Smooth Quantile Regression}

Zheng's smooth approximation \cite{zheng2011gradient} of the pinball function in Eq. \eqref{pinball} is
\begin{equation}
S_{\tau,\alpha}(u) = \tau u + \alpha \log \left( 1 + \exp \left(-\frac{u}{\alpha}\right) \right), 
\end{equation}

\noindent where  $ \alpha>0 $ is a smoothing parameter and $ \tau \in [0,1] $ is the quantile level we're  trying to estimate. 
Zheng proved in \cite{zheng2011gradient} that in the limit as $ \alpha \rightarrow 0^{+} $ that $ S_{\tau,\alpha}(u) = \rho_{\tau}(u) $. He also derives and discusses several other properties of the smooth pinball function. The smooth quantile regression optimization problem then becomes
\begin{equation} \label{Eq.smoothQR}
\min_{W,b} \Phi_{\tau,\alpha}(W,b) = \min_{W,b} \frac{1}{N} \sum_{t=1}^{N} S_{\tau,\alpha} (y_t-\hat{q}_{t}^{(\tau)} )
\end{equation}

\noindent where $ N $ is the number of training examples and $\hat{q}_{t}^{(\tau)} = W X_t + b$ where $W,b$ are the model parameters and $X_t$ is a vector of features at time $ t $. This form conveniently allows gradient based algorithms to be used for optimization.


%


\subsection{Neural Network Model and Optimization}

The smooth pinball neural network is designed for nonlinear multiple quantile regression. It has an input layer, output layer, and one hidden layer. The input layer consists of $ n_x $ number of input nodes and takes vector $ X_t $ of input features at time $ t $. The hidden layer consists of $ n_h $ number of hidden neurons and the output layer consists of $M$ number of output nodes corresponding to the estimated quantiles $ \hat{Q}_t = [\hat{q}_{t}^{(\tau_1)},...,\hat{q}_{t}^{(\tau_{M})}]^\top $ where $ \hat{q}_{t}^{(\tau_m)} $ is the $ \tau_m $ quantile level we want to estimate at time $t$. Every element in the first layer is connected to hidden neurons with the weight matrix $ W^{[1]} $ of size $ (n_x \times n_h) $ and bias vector $ b^{[1]} $of size $ (n_h \times 1) $. A similar connection structure is present in the second layer in the network between the hidden and output layers with $ W^{[2]} $ the output weight matrix of size $ (n_h \times M) $ and bias vector $ b^{[2]} $ of size $ (M \times 1) $. The input to hidden neurons is calculated, in vectorization notation, by
\begin{equation*}
Z_{t}^{[1]} =  W^{[1]} X_t + b^{[1]},
\end{equation*}

\noindent the output of the hidden layer then uses the logistic activation function
\begin{equation*}
H_{t} = \tanh \left(  Z_{t}^{[1]} \right). 
\end{equation*}

\noindent The input to output neurons is then calculated by
\begin{equation*}
Z_{t}^{[2]} =  W^{[2]} H_t + b^{[2]},
\end{equation*}

\noindent and the output layer uses the identity activation function
\begin{equation*}
\hat{Q}_t =  Z_{t}^{[2]}.
\end{equation*}


\noindent The objective function for our SPNN model is then the smooth pinball approximation summed over $ M $ number of $ \tau $'s we are trying to estimate in the output layer. We also use L2 regularization on the network weights in the objective function to prevent over-fitting during training. The objective function for SPNN is then given by 
\begin{equation} \label{opti}
\begin{multlined}
E = \frac{\lambda_1}{2NM}\| W^{[1]} \|^{2}_{F}+ \frac{\lambda_2}{2NM}\| W^{[2]} \|^{2}_{F}+  \frac{1}{NM}\sum_{t=1}^{N} \sum_{m = 1}^{M} ...\\
\left[ \tau_m (y_t - \hat{q}_{t}^{(\tau_m)}) + \alpha \log \left( 1 + \exp \left( - \frac{y_t-\hat{q}_{t}^{(\tau_m)}}{\alpha} \right) \right) \right]. 
\end{multlined}
\end{equation}

\noindent where $ \| . \|_{F} $ is the Frobenius norm. 

To train our model, we use standard gradient descent with backpropagation. Through this process we compute the gradient of the objective function $ E_t $ at each data point at time $ t $ with respect to $ W^{[1]}, b^{[1]}, W^{[2]} $ and $ b^{[2]} $. We start with the gradient with respect to the hidden-to-output weights $ W^{[2]}  $. In order to compute the gradient at time $ t $, we apply the chain rule in vector notation as follows
\begin{equation*}\label{wO}
\begin{multlined}
\frac{\partial E_t }{\partial W^{[2]}} = 
\frac{\lambda_2}{M} W^{[2]} + 
\frac{\partial E_t}{\partial \hat{Q}_{t}} \cdot \frac{\partial \hat{Q}_{t}}{\partial Z_{t}^{[2]}}
\cdot \frac{\partial Z_{t}^{[2]}}{\partial W^{[2]}}\\
=\frac{\lambda_2}{M} W^{[2]} +  \frac{1}{M} \left(\frac{1}{1+\exp \left( \frac{y_t-\hat{Q}_{t}}{\alpha} \right)} - T \right) H_{t}
\end{multlined},
\end{equation*}

\noindent where $ T = [\tau_1,...,\tau_m]^\top $ is a vector of all our $ \tau $'s. The gradient of $ b^{[2]} $ can be calculated similarly. Next we calculate the gradient of the objective function with respect to the weights of the first layer $ W_{[1]} $ as follows
\begin{equation*} \label{wI}
\begin{multlined}
\frac{\partial E_t}{\partial W_{[1]} } = \\ 
\frac{\lambda_1}{M} W^{[1]} + 
\left(  \frac{\partial E_t}{\partial \hat{Q}_{t}}  \cdot \frac{\partial \hat{Q}_{t}}{\partial Z_{t}^{[2]}}  \cdot \frac{\partial Z_{t}^{[2]}}{\partial H_{t} } \right) \cdot \frac{\partial H_{t}}{\partial Z_{t}^{[1]}} \cdot \frac{\partial Z_{t}^{[1]}}{\partial W^{[1]}}  \\
= \frac{\lambda_1}{M} W^{[1]} + \frac{1}{M}
\left( \frac{1}{1+\exp \left( \frac{y_t-\hat{Q}_{t}}{\alpha} \right)} - T \right) W^{[2]}\\
\left( 1-H_{t}^{2} \right) X_t
\end{multlined}.
\end{equation*}

\noindent The gradient of $ b^{[1]} $ can be calculated similarly. These gradients can then be directly used in gradient descent based optimization.

\subsection{Noncrossing Quantiles}

In quantile regression normally a single quantile is estimated. To estimate multiple quantiles the QR formulation could be run to solve for different $ \tau $'s independently. However in doing so quantiles may cross each other which is not desirable since it violates the principle of monotonically increasing inverse density functions. To prevent this, we need to introduce constraints as per \cite{takeuchi2006nonparametric}. The condition $ 0< \tau_1<...<\tau_M $ are defined as the orders of $ M $ conditional quantiles to be estimated. To ensure these quantiles do not cross each other the following constraint is needed $ q_{t}^{(\tau_1)} \leq ... \leq q_{t}^{(\tau_M)}, \forall_t $. However, it is not easy to solve the neural network optimization problem with such constraints using gradient descent methods.

To address this issue we developed a novel weight initialization scheme, similar to the mechanism of an extreme learning machine \cite{huang2006extreme}. Our scheme prevents quantiles from crossing over by initializing estimates of weights to fixed quantile values. In our approach we start by assuming all past observations of wind power correspond to a uniform distribution such that given a predictor $ X_t $ for all training observations $ t = 1 \dots N $ and $ M $ number of quantiles we're interested in estimating we get the following quantile output matrix
\begin{equation*}
Q = 
\begin{bmatrix}
	\tau_{1,1} & \dots & \tau_{1,M} \\
	\vdots & \ddots & \vdots \\
	\tau_{N,1} & \dots & \tau_{N,M} \\
\end{bmatrix}.
\end{equation*}

\begin{figure*}
	\begin{center}
		\includegraphics[width=.3\textwidth]{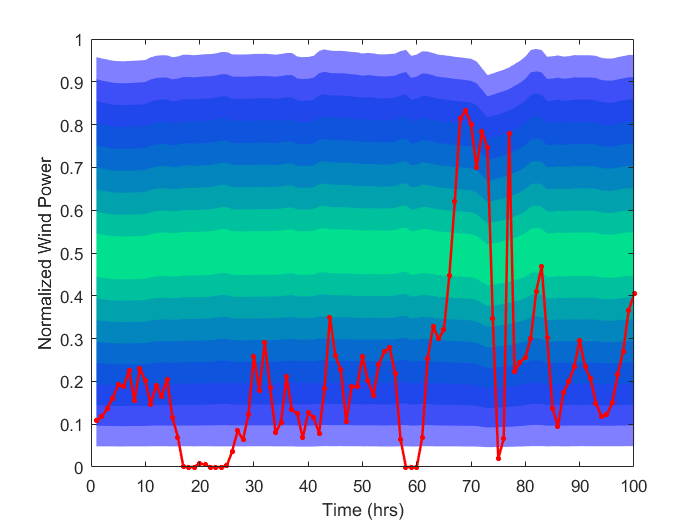}
		\includegraphics[width=.3\textwidth]{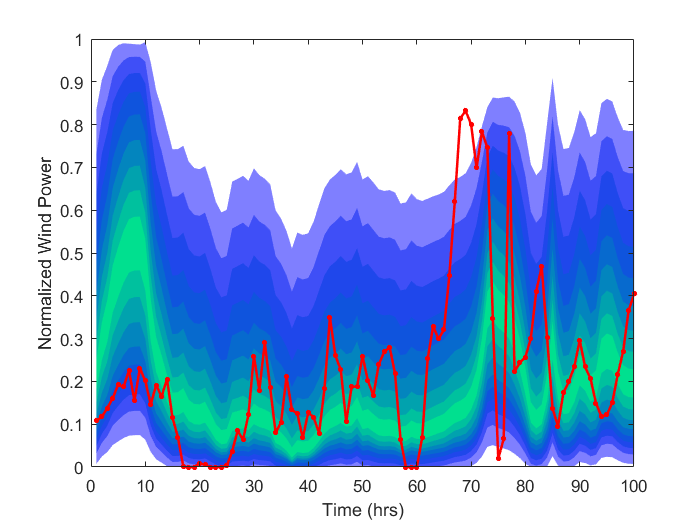}
		\includegraphics[width=.3\textwidth]{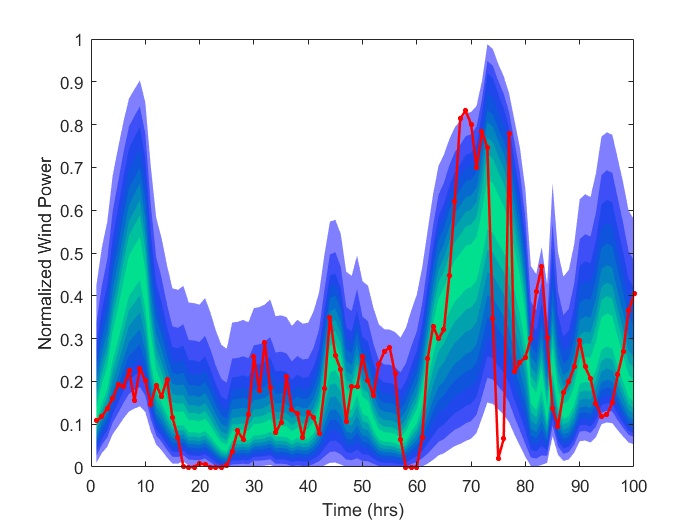}
	\end{center}
	\caption{Example training run showing growing quantiles out from the estimated mean with our weight initialization scheme at training iteration 1 (left), 100 (middle), and 1000 (right).}
	\label{iterations}
\end{figure*}

\noindent Our weight initialization then randomly assigns the input weights $ W^{[1]} $ of our neural network and sets the bias vectors $ b^{[1]}, b^{[2]} $ to zero. The output weights are then calculated as a least squared estimation of the above quantile outputs
\begin{equation*}
	W^{[2]} = H^{\dag} \cdot Q
\end{equation*} 

\noindent where $ H = \tanh(X \cdot W^{[1]}) $, $ X $ is a matrix of all training data, and $ H^{\dag} $ is the Moore-Penrose generalized inverse of matrix $ H $. Before training, all quantiles begin equidistant from each other. Then with each training iteration, the estimated quantiles move closer to a local optimum as reached by the objective function. Due to the symmetry of the weights during initialization, all estimated quantiles retain a certain proportional distance from each other as they move closer to better estimates. This scheme does not guarantee there are no cross-overs, but it does greatly reduce them as that it is seen in section \ref{s:res}. This process is illustrated in Fig. \ref{iterations} with an example training run for January 2013 from zone 1 showing the estimated quantiles after training iteration 1, 100, and 1000. Actual wind power is shown in red line for comparison.

\section{Case Study and Results}\label{s:res}

In validating our model for probabilistic forecasting of wind power we utilize wind data from the publicly available Global Energy Forecasting Competition 2014 (GEFCom2014) \cite{hong2016probabilistic}. The goal of the wind component of GEFCom2014 was to design parametric or nonparametric forecasting methods that would allow conditional predictive densities of the wind power generation to be described as a function of input data which were future weather forecasts and/or past wind data. Evaluation of predicted densities was done using the quantile score. Data is provided for the years of 2012 and 2013 from 10 wind farms titled Zone 1 to Zone 10. The predictors are numerical weather predictions (NWPs) in the form of wind speeds at an hourly resolution at two heights, 10m and 100m above ground level. These forecasts are for the zonal and meridional wind components (denoted U and V). It was up to the contestants to deduce exact wind speed, direction, and other wind features if necessary. These NWPs were provided for the exact locations of the wind farms. Additionally, power measurements at the various wind farms, with an hourly resolution, are also provided. All power measurements are normalized by the nominal capacity of their wind farm. The goal in forecasting is to learn to associate the provided NWPs (or derived features) with wind power. NWPs are provided for the forecasting horizon of one month and it is up to a forecasting model to use those NWPs as input to predict quantiles at each future time step. 

In our analysis of SPNN, we use the 12 months of 2013 from all 10 zones for testing. Training is done using a sliding window of two previous months to forecast the third month. For instance, to predict the output for June, training is done on observed data from April and May; then to predict output for July, training is done from May and June; and so forth. We derive twelve features from the raw data for training the SPNN model. Features used are derived wind speeds at 10m and 100m, wind direction at 10m and 100m, wind energy at 10m and 100m, hour of day, day of the year, and we also include in training the four raw wind speeds at 10m and 100m for U and V directions. All features were normalized using standardization. Denoting $ u $ and $ v $ as the wind components and $ d $ as the energy density (we use $ d=1 $), the equations used to compute the features are: wind speed $ =  \sqrt{u^2 + v^2} $, wind direction $ = \frac{180}{\pi} \times \arctan(u,v) $, and wind energy $  = \frac{1}{2} \times d \times \text{ws}^3 $.

We choose these twelve features as they are seen as the most commonly used across multiple studies examining probabilistic forecasting for wind farms \cite{hong2016probabilistic,zhang2014review}. More complex features could be used that also include correlated data from neighboring wind farms, time lagged features, wind shear, wind energy and direction differences between 10m and 100m. Additionally, if given exact location of the wind farms, one could extrapolate pressure, temperature, and other weather features. Most of the winning teams in the GEFCom2014 for the wind forecasting track used off-the-shelf machine learning or statistical methods such as multiple quantile regression for forecasting but conducted heavy manual feature selection to reduce the quantile score. The goal of our study is not custom feature engineering, which could result in better scores, but to showcase the feasibility of our method as a forecasting model given standard features and compare it with common and advanced benchmarks. 

\subsection{Benchmark Methods}

We use three standard \cite{sideratos2012probabilistic} and two advanced benchmark methods for density forecasting of wind power. The standard methods are the persistence model that corresponds to the normal distribution and is formed by the last 24 hours of observations, the climatology model that is based on all past wind power, and the uniform distribution that assumes all observations occur with equal probability. For our advanced benchmarks we then use multiple quantile regression (MQR) with L2 regularization, and support vector quantile regression (SVQR) \cite{hwang2005simple} with a radial basis function kernel.

\begin{table*}[t]
	\centering
	\caption{Evaluation scores from every zone averaged across every month.}
	\label{t:res}
	\begin{tabular*}{\textwidth}{@{\extracolsep{\fill}}rrrrrrrrr@{}}
		\toprule
		Score &	Zone & Uniform & Persistence & Climatology & MQR & SVQR & SPNN-wo & SPNN-w\\
		\midrule
		\multirow{10}{*}{QS} 
		&	1	&	0.1198	&	0.0834	&	0.0816	&	0.0520	&	0.0535	&	0.0498	&	\textbf{0.0491}	\\
		&	2	&	0.1065	&	0.0808	&	0.0796	&	0.0460	&	0.0465	&	0.0437	&	\textbf{0.0430}	\\
		&	3	&	0.1354	&	0.0899	&	0.0892	&	0.0446	&	0.0476	&	0.0431	&	\textbf{0.0418}	\\
		&	4	&	0.1378	&	0.0980	&	0.0953	&	0.0483	&	0.0492	&	0.0456	&	\textbf{0.0449}	\\
		&	5	&	0.1423	&	0.1012	&	0.0991	&	0.0500	&	0.0523	&	0.0478	&	\textbf{0.0468}	\\
		&	6	&	0.1405	&	0.1011	&	0.0991	&	0.0522	&	0.0532	&	0.0490	&	\textbf{0.0490}	\\
		&	7	&	0.1168	&	0.0803	&	0.0791	&	0.0392	&	0.0394	&	0.0365	&	\textbf{0.0364}	\\
		&	8	&	0.1187	&	0.0828	&	0.0816	&	0.0459	&	0.0470	&	0.0440	&	\textbf{0.0434}	\\
		&	9	&	0.1082	&	0.0804	&	0.0795	&	0.0420	&	0.0436	&	0.0401	&	\textbf{0.0397}	\\
		&	10	&	0.1502	&	0.1030	&	0.1010	&	0.0575	&	0.0576	&	0.0546	&	\textbf{0.0542}	\\
		\midrule	
		\multirow{10}{*}{IS} 
		&	1	&	-0.9580	&	-0.6673	&	-0.6530	&	-0.4160	&	-0.4280	&	-0.3983	&	\textbf{-0.3925}	\\
		&	2	&	-0.8523	&	-0.6465	&	-0.6364	&	-0.3679	&	-0.3723	&	-0.3512	&	\textbf{-0.3441}	\\
		&	3	&	-1.0833	&	-0.7195	&	-0.7132	&	-0.3565	&	-0.3811	&	-0.3442	&	\textbf{-0.3344}	\\
		&	4	&	-1.1027	&	-0.7842	&	-0.7627	&	-0.3858	&	-0.3933	&	-0.3650	&	\textbf{-0.3592}	\\
		&	5	&	-1.1382	&	-0.8097	&	-0.7930	&	-0.3997	&	-0.4180	&	-0.3820	&	\textbf{-0.3746}	\\
		&	6	&	-1.1243	&	-0.8089	&	-0.7924	&	-0.4173	&	-0.4256	&	-0.3919	&	\textbf{-0.3916}	\\
		&	7	&	-0.9342	&	-0.6422	&	-0.6325	&	-0.3136	&	-0.3840	&	-0.2920	&	\textbf{-0.2914}	\\
		&	8	&	-0.9495	&	-0.6626	&	-0.6532	&	-0.3674	&	-0.3757	&	-0.3518	&	\textbf{-0.3475}	\\
		&	9	&	-0.8658	&	-0.6432	&	-0.6359	&	-0.3362	&	-0.3488	&	-0.3205	&	\textbf{-0.3175}  	\\
		&	10	&	-1.2013	&	-0.8236	&	-0.8081	&	-0.4598	&	-0.4610	&	-0.4367	&	\textbf{-0.4362}	\\		
		\midrule	
		\multirow{10}{*}{ACE} 
		&	1	&	23.88	&	7.98	&	6.62	&	\textbf{3.26}	&	7.62	&	3.95			&	4.16			\\
		&	2	&	27.77	&	8.63	&	7.93	&	4.82			&	8.19	&	5.48			&	\textbf{4.80}	\\
		&	3	&	23.79	&	8.33	&	4.16	&	3.62			&	8.62	&	3.84			&	\textbf{3.60}	\\
		&	4	&	23.10	&	10.73	&	6.58	&	3.51			&	6.16	&	\textbf{2.93}	&	3.23			\\
		&	5	&	23.16	&	11.20	&	5.97	&	3.91			&	6.29	&	3.39			&	\textbf{3.37}	\\
		&	6	&	21.89	&	10.79	&	5.29	&	4.81			&	5.94	&	3.96			&	\textbf{3.55}	\\
		&	7	&	23.79	&	8.23	&	6.54	&	3.74			&	6.72	&	4.12			&	\textbf{3.40}	\\
		&	8	&	23.41	&	8.71	&	6.90	&	4.45			&	7.67	&	3.85			&	\textbf{3.82}	\\
		&	9	&	23.63	&	8.65	&	6.01	&	3.32			&	7.25	&	3.29			&	\textbf{2.96}	\\
		&	10	&	21.88	&	11.60	&	5.28	&	\textbf{3.78}	&	7.19	&	4.21			&	4.51			\\
		\bottomrule
	\end{tabular*}
\end{table*}

\subsection{Results and Discussion}

Our SPNN model is a fully connected feedforward neural network with one hidden layer and for weight optimization uses gradient descent backpropagation. The quality of the quantile estimates are sensitive to the hyperparameters of the network. SPNN has several hyperparameters that need to be chosen before training. The are the number of training iterations, number of hidden nodes, learning rate, smoothing rate, and L2 regularization terms. Through empirical testing on training data we found the following values as adequate for our model hyperparameters: 10000 training iterations, 20 hidden nodes, 0.3 for the learning rates, 0.01 for the smoothing rate, and 0.1 for each of the weight regularization terms.


\begin{figure}[t]
	\centering
	\includegraphics[width=0.5 \textwidth]{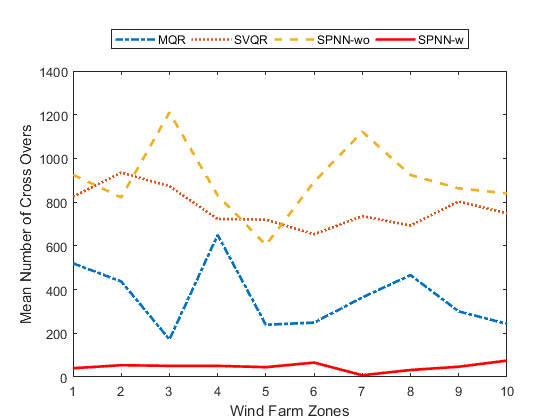}
	\caption{Mean number of quantile cross overs from each method across the 10 wind farm zones.}
	\label{crossOvers}
\end{figure}

We run our case study on a computer with an Intel i7 6700 2.6 GHz, and 16 GB of RAM. To ensure that our study is unbiased, we use for assessment the whole year of 2013. This gives a total of $ 365×\times24 = 8760 $ test samples for wind power forecasting per wind farm. With 10 wind farms in total, we have 87,600 test observations. Forecasting is conducted using a sliding window of the past 2 months for training to predict out 1 month. We apply our method along with benchmark methods to estimate 18 quantiles from which we also obtain 10\% to 90\% prediction intervals in increments of 10\%. Since for each test month we are estimating 18 quantiles for 720 look ahead hours across 10 farms, we need a way to best summarize so much data. Thus for every zone we take the mean of all the quantile estimates across all 12 months. Table \ref{t:res} summarizes results of QS, IS, and ACE scores averaged across all quantiles for each zone for the benchmark methods and the SPNN model without (wo) and with (w) our weight initialization scheme. 
Empirical analysis of the quantile estimates in Table \ref{t:res} shows that SPNN-w results in the lowest quantile scores across all ten zones by a significant amount. This provides evidence of the value of our proposed method in providing full predictive densities. We see that between SPNN-wo and SPNN-w there is a decrease in the drop of the QS meaning that smart weight initialization also leads to better performance

Inspecting the coverage analysis of our prediction intervals with the ACE score, we see that SPNN overall has the lowest or second lowest ACE for most of the zones. Looking at the sharpness of PIs with the interval score we see that, similar to the QS results, SPNN-w has the lowest across all zones. This means it produced the sharpest intervals. Since both QS and IS also both measure skill we can say that SPNN-w was able to produce the highest quality estimates from all methods. An interesting observation is the SPNN is designed to produce optimal quantile estimates and that indirectly it also produces adequate interval forecasts. If the main goal is to reduce ACE and IS as best as possible, alternative loss functions that incorporate prediction interval coverage and width functions
can be used. However, while not directly optimizing for coverage or sharpness, SPNN-wo and SPNN-w does yield results which are better from the advanced benchmarks multiple quantile regression and support vector quantile regression. Lastly we analyze the number of quantile cross overs counted when applying the methods to test data. Fig. \ref{crossOvers} showcases the averaged number of cross overs of the MQR, SVQR, and SPNN methods across the 10 wind farms. The uniform, persistence, and climatology models are omitted since by their nature the quantiles they estimate could never cross. We see in Fig. \ref{crossOvers} that cross overs for MQR and SVQR range from 200 to 900 for each zone and SPNN-wo is over 1000 for some zones. However, with the weight initialization scheme applied in SPNN-w we see a drop of almost two orders of magnitude in cross-overs.

With the results from Table \ref{t:res}, we have the following conclusions; demonstrated results imply the strong prediction ability and stability of the proposed prediction method. For the quantile cross over problem, for all zones and months, in every prediction we report few cross overs which validates our weighting scheme. Additionally, for all the runs across months and farms, the preassigned PI levels are satisfied which implies that the constructed PIs cover the target values with a high probability and with the lowest QS and IS we are able to obtain satisfactory predictive densities.

\section{Conclusion} \label{s:con}

Wind power forecasting is crucial for many decision making problems in power systems operations, and is a vital component in integrating more wind into the power grid. Due to the chaotic nature of the wind it is often difficult to forecast. Uncertainty analysis in the form of probabilistic wind prediction can provide a better picture of future wind coverage. This paper introduces a novel approach for probabilistic wind forecasting using a neural network with smooth approximation to the pinball ball loss function in estimating quantiles. We develop a novel weight initialization scheme where weights are fixed to a squared error approximation of the wind power corresponding to the uniform distribution, to ensure multiple quantiles can be estimated simultaneously without overlapping each other. We verify the effectiveness of our SPNN model  with the dataset of the Global Energy Forecasting Competition 2014. We compare forecasts  to common and advanced benchmarks and are evaluated using the quantile score, reliability, and sharpness metrics. Our results show superior performance across the prediction horizons, which verify effectiveness of the model for forecasting while preventing estimated quantiles from overlapping. 

Our SPNN method has the potential to be applied across a variety of domains for probabilistic forecasting or multiple quantile estimation. Future work will look into applying SPNN to forecast ocean wave and solar power, to test its effectiveness across different renewable energies, and on electricity pricing and load demand for smart grid applications. In this study we trained our model using NWP data. Another problem to study is very short term probabilistic forecasting using only past wind power data. Future work can also then look into expanding the SPNN model for providing full predictive densities given lagged past data of power only. New extensions need to be explored to ensure that past data is retained in memory by using deep or recurrent neural network architectures.

\bibliographystyle{IEEEtran}
\bibliography{mybib}

%
%
%

\end{document}